\documentclass[letterpaper]{article} 
\usepackage{aaai25}  
\usepackage{times}  
\usepackage{helvet}  
\usepackage{courier}  
\usepackage[hyphens]{url}  
\usepackage{graphicx} 
\usepackage{natbib}  
\usepackage{caption} 
\usepackage{algorithm}
\usepackage{algorithmic}
\usepackage{array}
\usepackage{amssymb}
\usepackage{amsmath}
\usepackage{bm}
\usepackage{booktabs}
\usepackage{adjustbox}
\usepackage{enumitem}
\usepackage{float}
\usepackage{lipsum}
\usepackage{subcaption}
\usepackage{multirow}
\usepackage{mathrsfs}
\usepackage{pdfpages}
\usepackage{placeins}

\usepackage{newfloat}
\usepackage{listings}
\DeclareCaptionStyle{ruled}{labelfont=normalfont,labelsep=colon,strut=off} 
\floatstyle{ruled}
\newfloat{listing}{tb}{lst}{}
\floatname{listing}{Listing}
\title{Bridge 2D-3D: Uncertainty-aware Hierarchical Registration Network with Domain Alignment}
\author {
    Zhixin Cheng\textsuperscript{\rm },
    Jiacheng Deng\textsuperscript{\rm },
    Xinjun Li\textsuperscript{\rm },
    Baoqun Yin\textsuperscript{\rm },
    Tianzhu Zhang\textsuperscript{\rm }
    \thanks{Corresponding author.}
}
\affiliations {
    \textsuperscript{\rm }Deep Space Exploration Laboratory/School of Information Science and Technology,\\
     University of Science and Technology of China\\
    {chengzhixin, dengjc, lxj3017}@mail.ustc.edu.cn,   {bqyin, tzzhang}@ustc.edu.cn

}

\usepackage{bibentry}

\begin{document}

\maketitle

\begin{abstract}
    
     The method for image-to-point cloud registration typically determines the rigid transformation using a coarse-to-fine pipeline. However, directly and uniformly matching image patches with point cloud patches may lead to focusing on incorrect noise patches during matching while ignoring key ones. Moreover, due to the significant differences between image and point cloud modalities, it may be challenging to bridge the domain gap without specific improvements in design. To address the above issues, we innovatively propose the Uncertainty-aware Hierarchical Matching Module (UHMM) and the Adversarial Modal Alignment Module (AMAM). Within the UHMM, we model the uncertainty of critical information in image patches and facilitate multi-level fusion interactions between image and point cloud features. In the AMAM, we design an adversarial approach to reduce the domain gap between image and point cloud. Extensive experiments and ablation studies on RGB-D Scene V2 and 7-Scenes benchmarks demonstrate the superiority of our method, making it a state-of-the-art approach for image-to-point cloud registration tasks.
\end{abstract}

\begin{figure}[!ht]
    \centering
    \begin{subfigure}[h]{\linewidth}
        \centering
        \includegraphics[width=\textwidth]{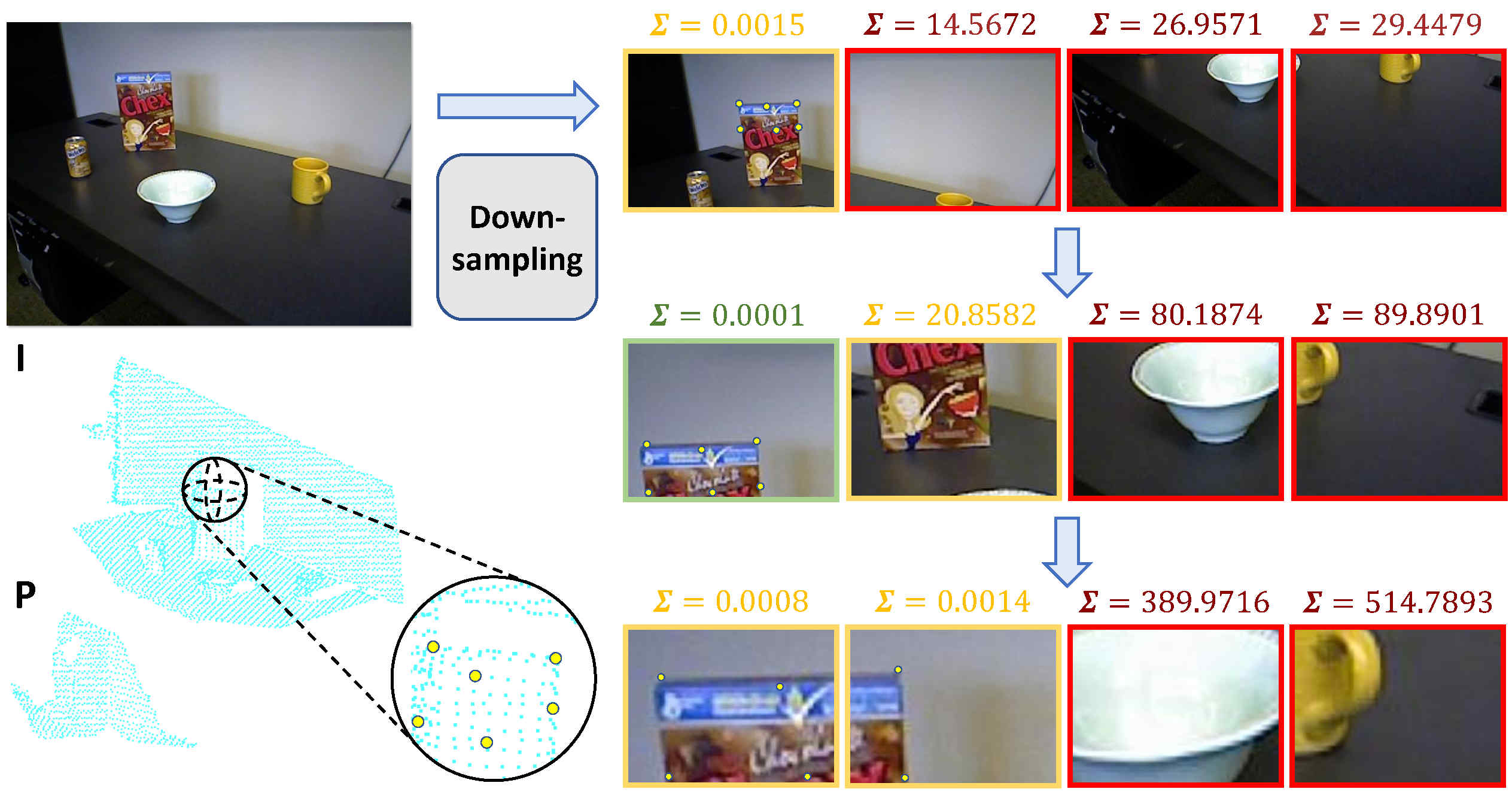} 
        \caption{}
        \label{fig:1a}
    \end{subfigure}
    
    \vspace{1em} 
    
    \begin{subfigure}[h]{\linewidth}
        \centering
        \includegraphics[width=\textwidth]{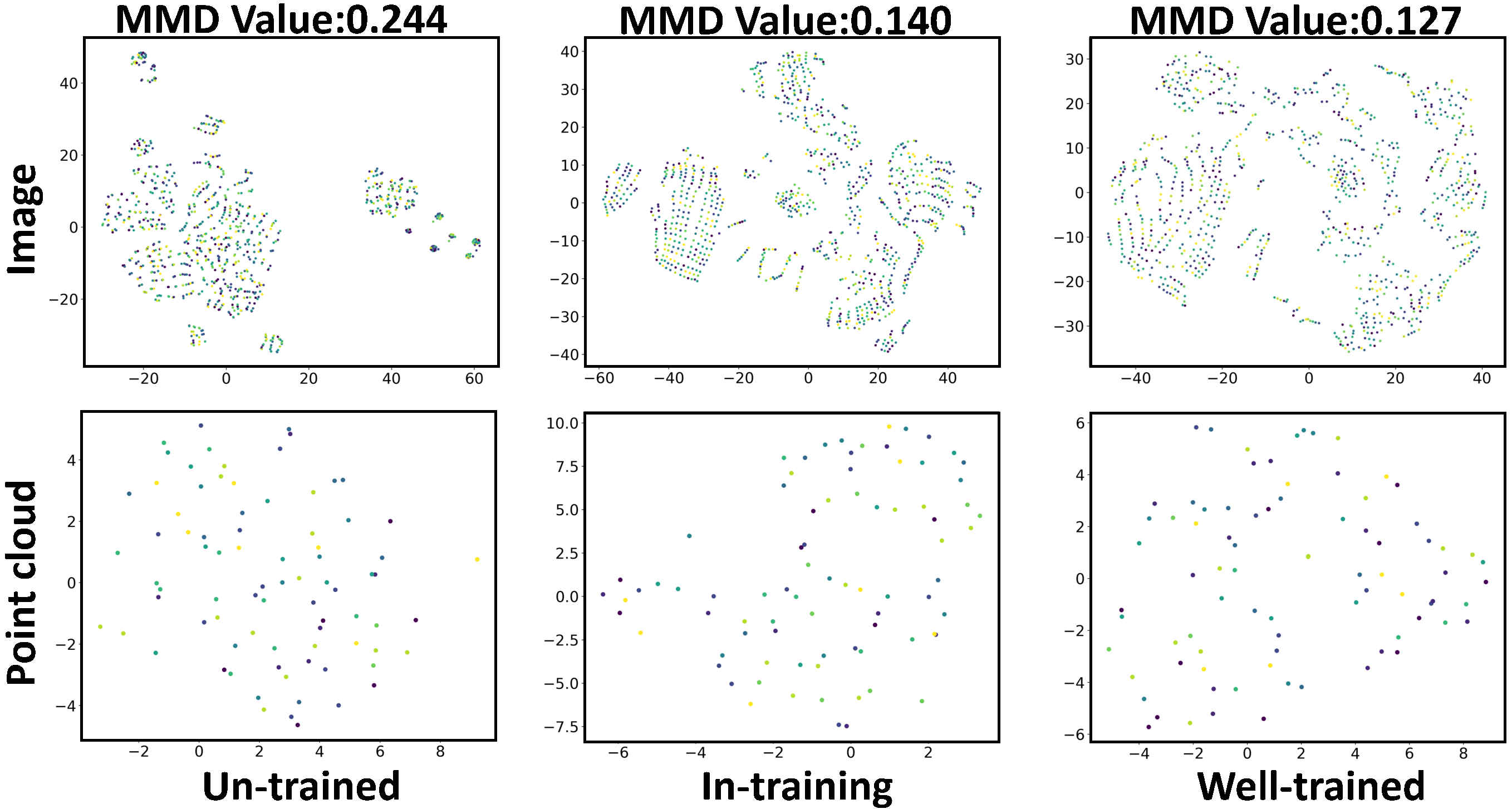} 
        \caption{}
        \label{fig:1b}
    \end{subfigure}

    \caption{
    (a) Visualization of uncertainty modeling, I and P form a pair. Variance indicates uncertainty: better matches have lower uncertainty, while poor matches have higher uncertainty.
    (b) Visualization of modality differences. With training, the modalities and distributions of the point cloud and image become aligned.
    }
    \label{fig:combined}
\end{figure}

\section{Introduction}
Image-to-point cloud registration aims to determine the rigid transformation from the point cloud to the camera coordinate system, which involves the cross-modal matching of image and point cloud, followed by a pose estimator to compute rotation and translation matrices. Such registration is crucial for 3D reconstruction \cite{3dreconstruction}, SLAM \cite{slam}, and visual localization \cite{visuallocalization} tasks. However, while images are dense 2D grids, point clouds are sparse and irregular 3D data. These fundamental differences pose challenges in cross-modal matching.

A range of methods has been proposed to address this challenge. Image-to-point cloud registration can be broadly categorized into detect-then-match \cite{2d3dmatchnet, p2, deepi2p, corri2p} and detection-free methods \cite{cofii2p, matr2d3d}. Detect-then-match methods first independently detect 2D key points in images and 3D key points in point clouds, then match them based on their semantic features. These methods face two main challenges: detecting key points is difficult because 2D key points rely on texture and color, while 3D key points depend on geometric structure. Additionally, 2D and 3D descriptors encode different visual information, complicating the extraction of consistent descriptors for matching. Thus, detection-free methods have gained prominence. The 2D3D-MATR \cite{matr2d3d} introduces a coarse-to-fine pipeline that first establishes patch-level matches between image and point cloud features, then refines these into dense pixel-to-point matches. Utilizing contextual information and different receptive fields significantly improves the inlier ratio. However, giving uniform attention to all image patches may lead to focusing on noisy patches and neglecting critical ones \cite{uncertainty}. Additionally, bridging the domain gap without specific design improvements \cite{domainunified} is challenging due to the substantial differences between image and point cloud modalities.

From the discussion of the detection-free method, two key issues need improvement for accurate and reliable matching between image and point cloud. Firstly, it is essential to differentiate and focus on vital image information during matching. While previous methods used patch-level matching to improve registration accuracy, emphasizing the importance of specific image patches can better address issues like receptive field misalignment and occlusion. Paying sufficient attention to crucial image patches during matching can enhance success rates and minimize the impact of noisy patches. As shown in Figure 1(a), the circular point patches correspond to the image patches on the right, displayed at different scales. They indicate potential matches with uncertainty levels increasing from green to yellow to red. As the scale decreases, uncertainty differences become more pronounced. Yellow dots mark vital points, and areas with more vital points have lower uncertainty. Secondly, reducing the differences between images and point clouds is crucial. The significant disparity between 2D and 3D data necessitates more consistent feature representation spaces to avoid incorrect matching of pixels and points. As illustrated in Figure 1(b), six t-SNE plots \cite{tsne} show images and point clouds at the un-trained, in-training, and well-trained stages, illustrating modality differences. The Maximum Mean Discrepancy (MMD) for each stage reflects domain changes due to the AMAM. Initially, images are dense, and point clouds are sparse, but after training, the modality differences lessen, and distributions become more unified. Designing a feature alignment module to ensure consistency of the acquired image and point cloud features is necessary, thereby improving registration efficiency, robustness, and generalization in subsequent tasks.

Inspired by the above discussions, we propose an uncertainty-aware hierarchical registration network named B2-3Dnet, including two innovative modules: the Uncertainty-aware Hierarchical Matching Module (UHMM) and the Adversarial Modal Alignment Module (AMAM). In the UHMM, we model uncertainty within image patches and perform multi-level fusion between image and point cloud features. We predict the mean and variance of image features at different scales and then resample image patch features. By constraining the sum of variances of all image patches through adaptive variance allocation, the network optimizes critical image patches while suppressing noise. After similarity calculation, we hierarchically fuse and interact multiple scale image patch features with point patch features from coarse to fine, allowing point cloud features to perceive image features with different receptive fields for subsequent similarity computation. In the AMAM, we design a domain adaptation method based on adversarial training to mitigate feature space differences between images and point clouds. This module includes a Gradient Reversal Layer (GRL) \cite{grl} and a domain classifier. The domain classifier determines whether the input feature comes from the image or point cloud domain and is constrained by classification loss. The classification loss gradient undergoes reversal when back-propagated through the GRL layer, promoting the extraction of features with minimal modal differences in the feature space.

In summary, our work can be summarized as follows:
\begin{itemize}
    \item We propose the B2-3Dnet, a novel uncertainty-aware hierarchical registration network with domain alignment, demonstrating excellent accuracy and strong generalization in image-to-point cloud registration tasks.
    \item We design the UHMM, employing uncertainty modeling of image patch importance and achieving multi-level feature interaction between image and point cloud patches. Additionally, we propose AMAM to address modal differences between images and point clouds.
    \item Our method has been extensively tested on two benchmarks, RGB-D Scene V2 \cite{rgbdv2} and 7-Scenes \cite{7scenes}, proving its superiority and establishing it as a state-of-the-art method in image-to-point cloud registration tasks.
\end{itemize}

\section{Related Work}

In this section, we briefly overview related works on image-to-point cloud registration, including stereo image registration, point cloud registration, and inter-modality registration. 

\textbf{Stereo Image Registration.} Detector-based methods have long dominated stereo image registration. Prior to deep learning, key points were detected using handcrafted techniques like SIFT \cite{sift} and ORB \cite{orb}, which built 2D matches from local features. The advent of deep learning introduced neural network-based detection, transforming the field. SuperGlue \cite{superglue} was pioneering in using Transformers \cite{transformer} for image registration, greatly enhancing local feature matching. However, the challenge of detecting repeatable interest points in non-salient areas has led to the rise of detector-free methods. Approaches like LoFTR \cite{loftr} and Efficient LoFTR \cite{efficientloftr} use a coarse-to-fine pipeline with Transformers to efficiently estimate dense image matches through global receptive fields.

\textbf{Point Cloud Registration.} Point cloud registration has advanced from handcrafted descriptors like PPF \cite{ppf} and FPFH \cite{fpfh} to deep learning-based approaches. CoFiNet \cite{cofinet} introduced detector-free registration with a coarse-to-fine strategy, and recent methods have replaced RANSAC \cite{ransac} with deep robust estimators for better speed and accuracy. GeoTransformer \cite{geotransformer} improves inlier ratios by integrating global information with the transformer and introduces a local-to-global method for RANSAC-free registration.

\textbf{Inter-modality Registration.} Inter-modal registration presents greater challenges compared to intra-modal registration because of the significant domain discrepancies involved. Traditional approaches typically employ a detect-then-match strategy. For instance, 2D3D-Matchnet \cite{2d3dmatchnet} uses SIFT \cite{sift} and ISS \cite{iss} to extract key points from images and point clouds, constructing patches around these key points. It then uses CNNs and PointNet \cite{pointnet} to extract features and build descriptors for matching. P2-Net \cite{p2} introduces a joint learning framework with a comprehensive reception mechanism, using a single forward pass to detect key locations and extract descriptors, enabling efficient matching through contrastive constraints. Unfortunately, the inefficiency of keypoint extraction in cross-modal scenarios has led to significant accuracy loss, prompting the emergence of detection-free methods. 2D3D-MATR \cite{matr2d3d} adopts a coarse-to-fine matching process, using a transformer-based approach to establish patch-level matches and then seeking fine-grained matches within them, followed by regressing rigid transformations using PNP+RANSAC \cite{pnp, ransac}. This detection-free method overcomes the challenge of obtaining repeatable key points and makes 2D-3D descriptors more consistent. Using transformers’ global receptive fields and multi-level methods significantly increases the inlier ratio of matches. 
Our proposed B2-3Dnet follows the detection-free approach, modeling uncertainty in the coarse matching process and effectively bridging the modality gap between images and point clouds. These innovations make our method state-of-the-art in image-to-point cloud registration.

\begin{figure*}[ht]
    \centering
    \includegraphics[width=\textwidth]{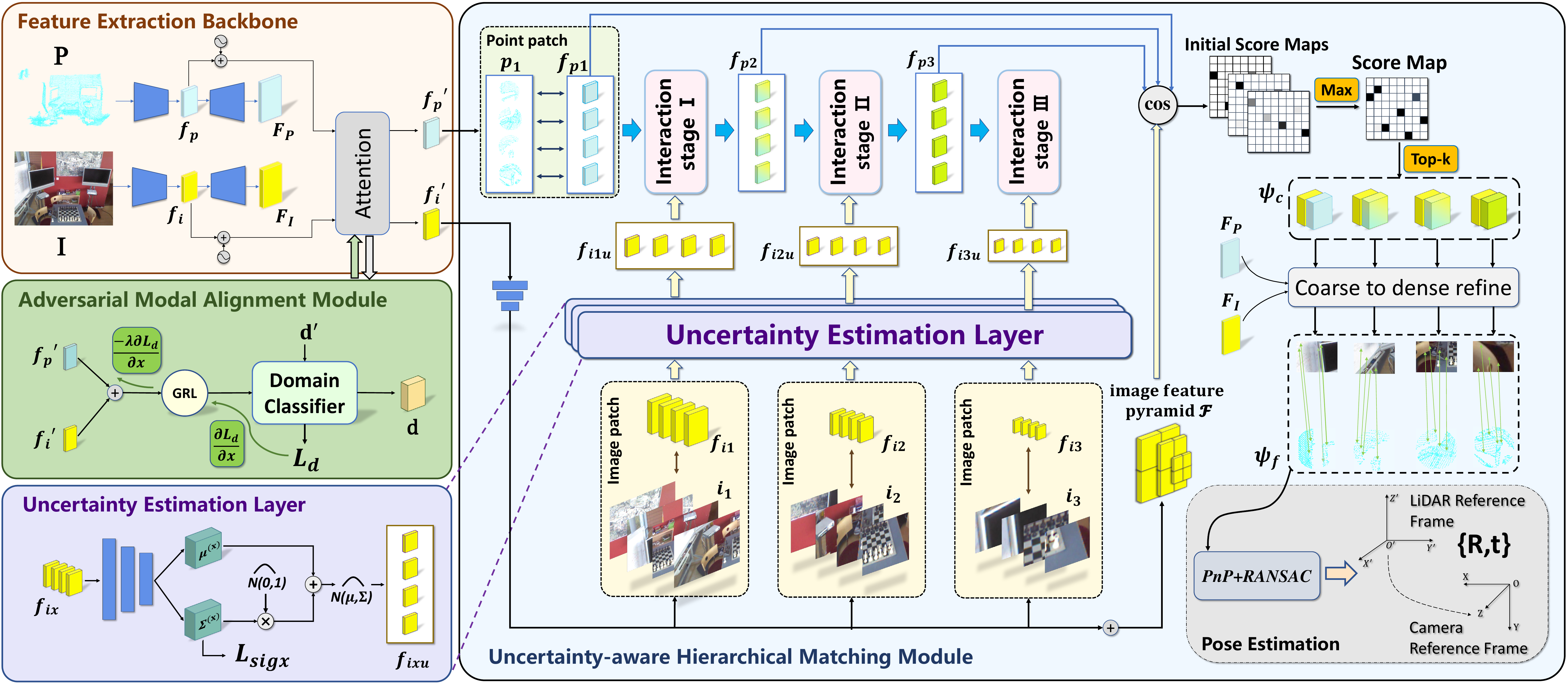} 
    \caption{Overall pipeline of B2-3Dnet. We use a feature extraction backbone to obtain features from images and point clouds, which are aligned using the adversarial modal alignment module to reduce domain differences. The image features are processed through hierarchical layers and uncertainty estimation layers to create informative image patches. During the interaction stages, updated point cloud patches and image features generate a score map via cosine similarity and maximum, achieving coarse-level matching and refining fine-level matches. Finally, PnP+RANSAC is used to regress the rigid transformation.}
    \label{fig:2}
\end{figure*}

\section{Method}

\subsection{Overview} 
Given an image \(\mathbf{I} \in \mathbb{R}^{W \times H \times 3}\) and a point cloud \(\mathbf{P} \in \mathbb{R}^{N \times 3}\) from the same scene, the objective of the registration between images and point clouds is to determine the rigid transformation \([R, t]\) from the point cloud coordinate system to the camera coordinate system. Here, \(W\) and \(H\) are the image's width and height, while \(N\) is the number of points. The transformation consists of a 3D rotation \(R \in \mathit{SO}(3)\) and a 3D translation vector \(t \in \mathbb{R}^3\).

Our method, B2-3Dnet, depicted in Figure 2, adopts a hierarchical structure comprising four main components: the feature extraction backbone, the Uncertainty-aware Hierarchical Matching Module (UHMM), the Adversarial Modal Alignment Module (AMAM), and the subsequent pose estimation. Initially, we extract features from images and point clouds using different feature extractors. These features are aligned via the AMAM before being processed by the UHMM. The UHMM applies uncertainty modeling to achieve coarse patch-level matches, which are then refined into dense matches. Finally, the PnP-RANSAC algorithm effectively computes the rigid transformation.

\subsection{Feature Extraction Backbone}
We use ResNet \cite{resnet} with FPN \cite{fpn} for image feature extraction and KPFCNN \cite{kpfcnn} for point cloud features. The 2D and 3D features are downsampled to \( f_i \in \mathbb{R}^{h \times w \times c} \) and \( f_p \in \mathbb{R}^{n \times c} \) for coarse matching. At the original resolution, they are represented as \( \mathbf{F}_I \in \mathbb{R}^{H \times W \times C} \) and \( \mathbf{F}_P \in \mathbb{R}^{N \times C} \) for fine matching. Positional encoding enhances \( f_i \) and \( f_p \), which are refined through self- and cross-attention in transformers, yielding improved features \( {f_i}' \) and \( {f_p}' \).

\subsection{Uncertainty-aware Hierarchical Matching Module}
After the feature extraction backbone, including a series of attention layers, we obtain the image and point cloud features \({f_i}'\) and \({f_p}'\). To achieve coarse matching in patches, the image \(\mathbf{I} \in \mathbb{R}^{H \times W}\) is divided into \(h \times w\) patches. A lightweight three-stage CNN \cite{lightweightcnn} is used to extract image patches at three different scales, denoted as \([i_1, f_{i1}]\), \([i_2, f_{i2}]\), and \([i_3, f_{i3}]\).

 In point cloud registration, a point-to-node partition \cite{pointtonode} maps points in \(\mathbf{P} \in \mathbb{R}^N\) to nearest grid points in \(\mathbf{p} \in \mathbb{R}^n\), forming point patches \([p_1, f_{p1}]\). A multi-level approach addresses scale ambiguity due to perspective effects, where nearby objects in images appear larger, and distant ones smaller, while the point cloud maintains a consistent scale. Hierarchical features ensure accurate matching at the correct scale, avoiding misalignment.

\textbf{Uncertainty estimation layer.}
Although the hierarchical design effectively builds multi-scale image features for coarse matching with point clouds, it is crucial to distinguish the importance of image patch information to address issues such as receptive field misalignment and occlusion from 2D and 3D feature extraction. Image patches with higher significance should receive more attention to enhance the matching success rate and minimize the impact of noisy patches. To achieve this, we designed an uncertainty estimation layer to assess image patch importance through uncertainty estimation, enabling differential treatment of image patches.

In the uncertainty estimation layer, we reconstruct features of image patches at different scales. The input features \(f_{ix}\), where \(x \in \{1, 2, 3\}\), are fed into three uncertainty estimation layers. We model the feature distribution of a single image as a Gaussian distribution. The features \(f_{ixu}\) obtained through uncertainty modeling for the \(x\)-th layer are reconstructed from a Gaussian distribution parameterized by the mean vector \(\mu^{(x)}\) and covariance matrix \(\Sigma^{(x)}\), predicted by the network. This process is illustrated in the uncertainty estimation layer area of Figure 2.

We utilize a term \( q \), representing entropy, where the variance is positively correlated, primarily to prevent the variance of flat solutions from approaching zero. The term \( q \) effectively helps maintain the level of uncertainty in the training samples and is defined as follows:
\begin{equation}
q = \frac{1}{2} \log\bigl(\det(2\pi e \Sigma)\bigr),
\end{equation}
the larger the variance, the greater the entropy. Through computation, the total uncertainty loss formula is derived by combining all the layers' losses as follows:
\begin{equation}
L_{\text{sig}} = \max(0, \gamma - \sum_{x=1}^{3} q^{(x)}),
\end{equation}
where \(\gamma\) is the threshold value for the total uncertainty sum, and \( q^{(x)} \), where \( x \in \{1, 2, 3\} \), represents the entropy of each layer. Evidently, utilizing \(L_{\text{sig}}\), the model aims to maintain the overall level of variance of the training samples.

Here, we emphasize the use of the reparameterization trick in sampling. Direct sampling produces random feature samples that do not allow gradients to propagate back to previous layers. To address this, we first draw a sample \(\bm{\varepsilon}\) from a standard Gaussian distribution with zero mean and unit covariance, \(\bm{\varepsilon} \sim \mathcal{N}(0, I)\). We then compute the sample using \(\mu + \bm{\varepsilon} \Sigma\) instead of directly drawing from \(\mathcal{N}(\mu, \Sigma)\). This approach decouples the sampling process from parameter training, enabling effective backpropagation of gradients.

Through \(L_{\text{sig}}\), our network gains\textbf{ two key capabilities}: 
 1. It assigns smaller variances to correctly matched image patches and larger variances to incorrectly matched ones.
 2. Image patches with larger variances have less impact on model training, minimizing the effect of incorrect matches that might otherwise misguide the network.

We will analyze these reasons in detail in the discussion section. By integrating these capabilities, the network is empowered to focus more effectively on extracting critical image patches. This targeted approach facilitates a more accurate alignment of point cloud patches with the most critical regions of the image. As a result, the overall matching process becomes more precise and reduces the impact of less relevant or noisy patches.

\textbf{Interaction stage.}
After obtaining the multi-scale image features \(f_{i1u}\), \(f_{i2u}\), and \(f_{i3u}\) through uncertainty modeling, we do not immediately perform coarse-level matches like the previous methods. Instead, we process them with point cloud patches during the interaction stage. Hierarchical features enable better scale-specific matching and help avoid feature misalignment. However, direct multi-scale matching may not be optimal due to the differences between images and point clouds and the untapped contextual information across image scales. To address this, we use cross-attention to facilitate interaction between image and point cloud patches. We then calculate the cosine similarity between the updated point cloud patches and the image feature pyramid across hierarchical patches to generate the initial score maps. For the first layer, we compute the query, key, and value matrices as follows:
\begin{equation}
Q = f_{p1} \cdot W_q, \quad K = f_{i1u} \cdot W_k, \quad V = f_{i1u} \cdot W_v,
\end{equation}
the matrices \( W_q \), \( W_k \), and \( W_v \) are learnable weight matrices for query, key, and value transformations, respectively. The attention matrix \( A \) is calculated using the following formula:
\begin{equation}
A = \text{softmax}\left(\frac{Q \cdot K^T}{\sqrt{d_k}}\right),
\end{equation}
where \( K^T \) represents the transposed key matrix, and \( d_k \) is the dimensionality of the key vectors used for scaling. The attention features are further projected with a shallow MLP as the final output features \(f_{p2}\). Following a similar process, we update to obtain \( f_{p3} \).

We construct an image feature pyramid $\mathcal{F} = \{f_{1}, f_{2}, f_{3}\}$ through concatenation. After normalizing this pyramid to the same dimension, we compute the cosine similarity \cite{cos} between the image feature pyramid and the point cloud features \( f_{p1}, f_{p2}, f_{p3} \), yielding in three initial score maps.

To create the final Score Map, we take the maximum value at each corresponding position across the three Initial Score Maps. During inference, patch-level matches \( \psi_c \) are extracted through mutual top-\( k \) \cite{topk} selection in the score map. Then, we refine the local region's pixels and points back to the initial resolution in the feature images (\(\mathbf{F}_I\)) and feature point clouds (\(\mathbf{F}_P\)). We then normalize and again perform top-\(k\) selection to achieve dense matches \(\psi_f\). Since 2D patches of different scales can overlap, we remove duplicate correspondences from the final match results.

\textbf{Discussion.} \textit{Why do we only model uncertainty for image patches?} Modeling uncertainty for both image and point cloud patches is computationally expensive and hard to converge. Therefore, we choose to model uncertainty only for images. Images, being two-dimensional and arranged in regular grids, are more straightforward to model compared to unordered, sparse, and irregular three-dimensional point clouds. Multi-level image patches naturally incorporate multi-scale information, effectively conveying contextual details for more accurate and robust matches. Our experiments confirm this approach's effectiveness.

\textit{Why does the network exhibit these two capabilities?} First, let us examine the loss functions for the coarse and fine-matching networks. Both \(\mathcal{L}_{\text{coarse}}\) and \(\mathcal{L}_{\text{fine}}\) use the general circle loss \cite{circleloss,circleloss1}. Given an anchor descriptor \(d_i\), the descriptors of its positive and negative pairs are \(\mathcal{D}^P_i\) and \(\mathcal{D}^N_i\), respectively:

\begin{equation}
\begin{aligned}
\mathcal{L}_i = \frac{1}{\gamma} \log\biggl[1 + \Bigl(&\sum_{d^j \in \mathcal{D}^P_i} e^{\beta^{i,j}_p(d^j_i - \Delta_p)}\Bigr) \\
&\cdot \Bigl(\sum_{d^k \in \mathcal{D}^N_i} e^{\beta^{i,k}_n(\Delta_n - d^k_i)}\Bigr)\biggr],
\end{aligned}
\end{equation}

where \(d^j_i\) is the \(L_2\) feature distance, \(\beta^{i,j}_p = \gamma \lambda^{i,j}_p(d^j_i - \Delta_p)\) and \(\beta^{i,k}_n = \gamma \lambda^{i,k}_n(\Delta_n - d^k_i)\) are the individual weights for the positive and negative pairs, with \(\lambda^{i,j}_p\) and \(\lambda^{i,k}_n\) as scaling factors \cite{overlap}. Since fine-level matching derives from coarse-level matches, uncertainty modeling mainly affects \(\mathcal{L}_{\text{coarse}}\). With this understanding, let us discuss these two capabilities of the model.

\begin{figure}[ht]
    \centering
    \small
    \includegraphics[width=\linewidth]{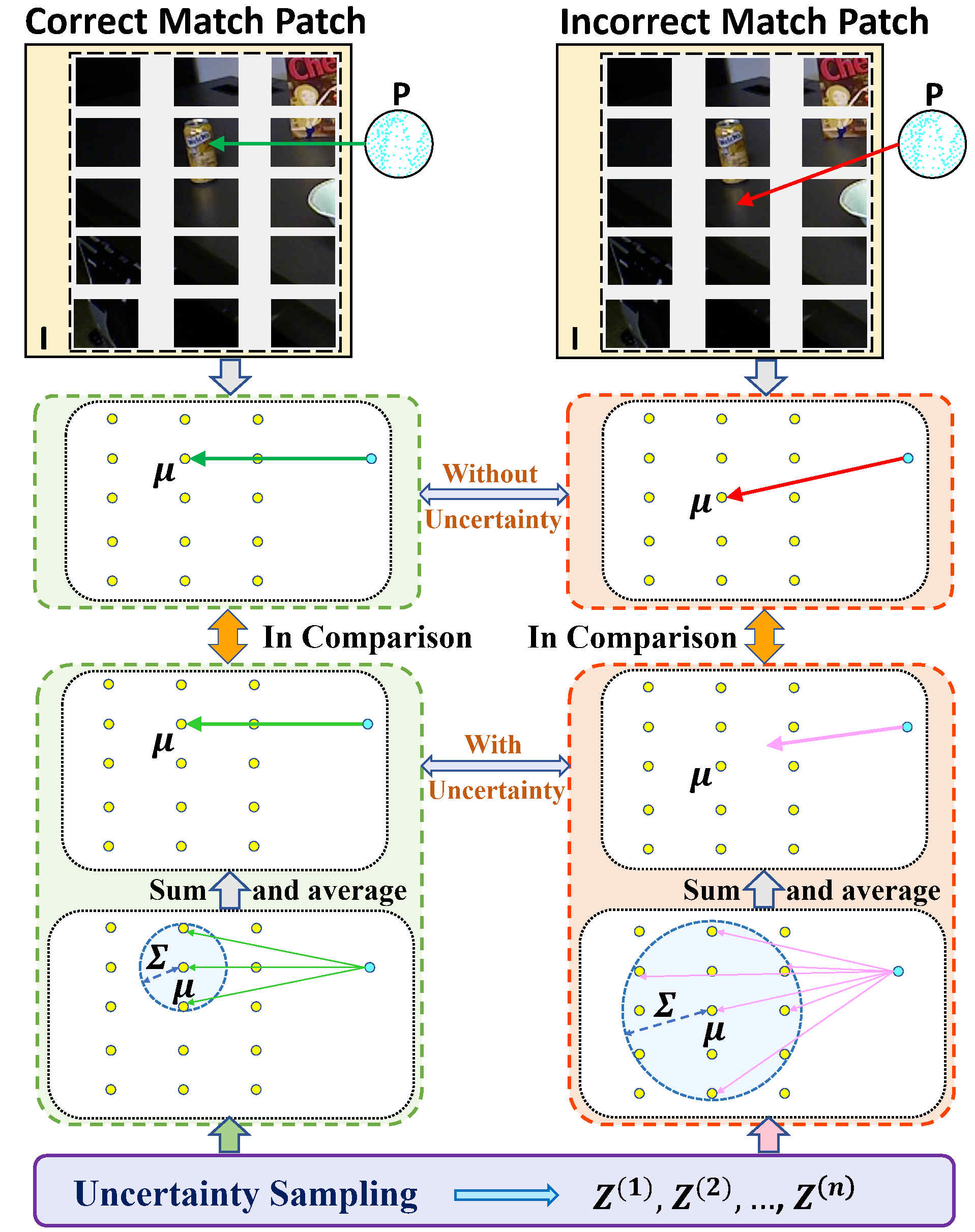} 
    \caption{Comparison of the effects with or without an uncertainty estimation layer on correctly or incorrectly matched patches.}
    \label{fig:3}
\end{figure}

\textit{Why is larger variance assigned to misaligned image patches?} 
The circle loss constrains \(\mathcal{L}_{\text{coarse}}\), where incorrect matches increase \(d^j_i\), reducing the difference between positive and negative samples, which in turn raises \(\mathcal{L}_{\text{coarse}}\). Image patches with larger variances result in higher \(\mathcal{L}_{\text{coarse}}\) due to the larger sampling space and more incorrect matches. However, we control the total variance and cannot satisfy \(\mathcal{L}_{\text{coarse}}\) by reducing all variances to zero. So, which image patches should be assigned a larger variance? Misaligned image patches will still cause a large \(\mathcal{L}_{\text{coarse}}\) even with smaller variance, but reducing the variance of correctly matched patches can directly reduce \(\mathcal{L}_{\text{coarse}}\). Therefore, the model assigns larger variance to misaligned image patches.

\textit{Why do image patches with larger variance have less impact on model training?} If an image patch is assigned a large uncertainty, the samples from this patch will have a large variance \(\Sigma\), causing the result \(Z\) to deviate from the original image patch \(\mu\). Consequently, when collecting feature space vectors between point cloud patches to different \(Z^{(1)}, Z^{(2)}, \ldots, Z^{(n)}\) and averaging them, their gradients may cancel each other out, as illustrated in Figure 3. Conversely, when the variance of an image patch is small, consistent gradients will be generated after entering the network, thereby enhancing its importance. Thus, our uncertainty modeling can determine which image patches are more or less important, altering their influence on model training to better focus on key information patches and reduce the impact of noisy patches.

\subsection{Adversarial Modal Alignment Module}

Images are two-dimensional data represented on a regular, dense grid at a specific resolution, while point clouds are unordered, sparse, and irregular three-dimensional data. Feature extractors are designed based on their different characteristics, resulting in significant domain differences between image and point cloud features, which become a major obstacle in the matching process. To address this, we designed an adversarial modal alignment module using adversarial concepts to align the image and point cloud modalities for better matching.

We label the extracted features \(f'_i\) from images and \(f'_p\) from point clouds with \(d'\), distinguishing between image (denote as label 1) and point cloud (denote as label 0) domains. These features are then fed into a domain classifier, composed of three connected layers, to predict the label \(d\). The cross-entropy loss \cite{crossentropy} between the predicted label and the true domain label is calculated to obtain \(L_d\). The domain alignment loss, \(L_d\), is defined as follows:
\begin{equation}
L_d = -\frac{1}{N} \sum_{n=1}^{N} \left[ d'_n \log(d_n) + (1 - d'_n) \log(1 - d_n) \right],
\end{equation}
where \(N\) is the total number of samples, \(d'_n\) is the true label for the \(n\)-th sample, and \(d_n\) is the predicted value for the \(n\)-th sample. 

The gradient of \(L_d\) is backpropagated as \(\frac{\partial L_d}{\partial x}\), and after passing through the Gradient Reversal Layer (GRL), it is inverted to \(-\lambda \frac{\partial L_d}{\partial x}\) before propagating back to the feature extraction backbone network. The inverted gradient represents the differences in feature space between image and point cloud domains learned by the classifier. By inverting these gradients, the differences are mitigated, aligning the extracted features more closely. This aligned feature is then classified, ensuring it retains its intrinsic characteristics despite modal alignment. With this adversarial approach, the feature extractor is optimized, obfuscating classification by domain, ultimately achieving alignment of image and point cloud feature domains.

\subsection{Model Training \& Inference}

\textbf{Training.} Incorporating the four types of losses mentioned earlier, we use \(\mathcal{L}_{\text{coarse}}\) and \(\mathcal{L}_{\text{fine}}\) to constrain the coarse and fine-level matching relationships. \(\mathcal{L}_{\text{sig}}\) focuses on key image patches by limiting the total variance, while \(\mathcal{L}_{d}\) promotes alignment of image and point cloud features by reversing gradients during backpropagation. Thus, our total loss function is expressed as:
\begin{equation}
\mathcal{L} = \mathcal{L}_{\text{coarse}} + \mathcal{L}_{\text{fine}} + \mathcal{L}_{\text{sig}} + \mathcal{L}_{d}.
\end{equation}
\textbf{Inference.} During inference, the computations from the adversarial modal alignment module are not required, ensuring the efficiency of the inference process.

\section{Experiments}

\subsection{Datasets and Implementation Details}
Based on the 2D3D-MATR benchmark, we conducted extensive experiments and ablation studies on two challenging benchmarks: RGB-D Scenes V2 \cite{rgbdv2} and 7Scenes \cite{7scenes}.

\begin{table}[ht]
\centering
\small
\setlength{\tabcolsep}{1mm} 
\resizebox{\columnwidth}{!}{ 
\begin{tabular}{@{}l@{\hskip 1pt}|ccccc@{}}
\toprule
\multicolumn{1}{l|}{Model} & Scene.11 & Scene.12 & Scene.13 & Scene.14 & Mean \\ \midrule
\multicolumn{1}{l|}{Mean depth (m)} & 1.74 & 1.66 & 1.18 & 1.39 & 1.49 \\ \midrule
\multicolumn{6}{c}{\textit{Inlier Ratio} ↑} \\ \midrule
\multicolumn{1}{l|}{FCGF-2D3D} & 6.8 & 8.5 & 11.8 & 5.4 & 8.1 \\
\multicolumn{1}{l|}{P2-Net} & 9.7 & 12.8 & 17.0 & 9.3 & 12.2 \\
\multicolumn{1}{l|}{Predator-2D3D} & 17.7 & 19.4 & 17.2 & 8.4 & 15.7 \\
\multicolumn{1}{l|}{2D3D-MATR} & \underline{32.8} & \textbf{34.4} & \underline{39.2} & \underline{23.3} & \underline{32.4} \\
\multicolumn{1}{l|}{B2-3Dnet (ours)} & \textbf{36.4} & \underline{32.7} & \textbf{43.8} & \textbf{27.4} & \textbf{35.1} \\ \midrule
\multicolumn{6}{c}{\textit{Feature Matching Recall} ↑} \\ \midrule
\multicolumn{1}{l|}{FCGF-2D3D} & 11.1 & 30.4 & 51.5 & 15.5 & 27.1 \\
\multicolumn{1}{l|}{P2-Net} & 48.6 & 65.7 & 82.5 & 41.6 & 59.6 \\
\multicolumn{1}{l|}{Predator-2D3D} & 86.1 & 89.2 & 63.9 & 24.3 & 65.9 \\
\multicolumn{1}{l|}{2D3D-MATR} & \underline{98.6} & \underline{98.0} & \underline{88.7} & \underline{77.9} & \underline{90.8} \\
\multicolumn{1}{l|}{B2-3Dnet (ours)} & \textbf{100.0} & \textbf{99.0} & \textbf{92.8} & \textbf{85.8} & \textbf{94.4} \\ \midrule
\multicolumn{6}{c}{\textit{Registration Recall} ↑} \\ \midrule
\multicolumn{1}{l|}{FCGF-2D3D} & 26.5 & 41.2 & 37.1 & 16.8 & 30.4 \\
\multicolumn{1}{l|}{P2-Net} & 40.3 & 40.2 & 41.2 & 31.9 & 38.4 \\
\multicolumn{1}{l|}{Predator-2D3D} & 44.4 & 41.2 & 21.6 & 13.7 & 30.2 \\
\multicolumn{1}{l|}{2D3D-MATR} & \textbf{63.9} & \underline{53.9} & \underline{58.8} & \underline{49.1} & \underline{56.4} \\
\multicolumn{1}{l|}{B2-3Dnet (ours)} & \underline{58.3} & \textbf{60.8} & \textbf{74.2} & \textbf{60.2} & \textbf{63.4} \\ \bottomrule
\end{tabular}
}
\caption{Evaluation results on RGB-D Scenes V2. \textbf{Boldfaced} numbers highlight the best and the second best are \underline{underlined}.}
\label{tab:evaluation}
\end{table}

\begin{figure*}[t]
\centering
\includegraphics[width=1.0\textwidth]{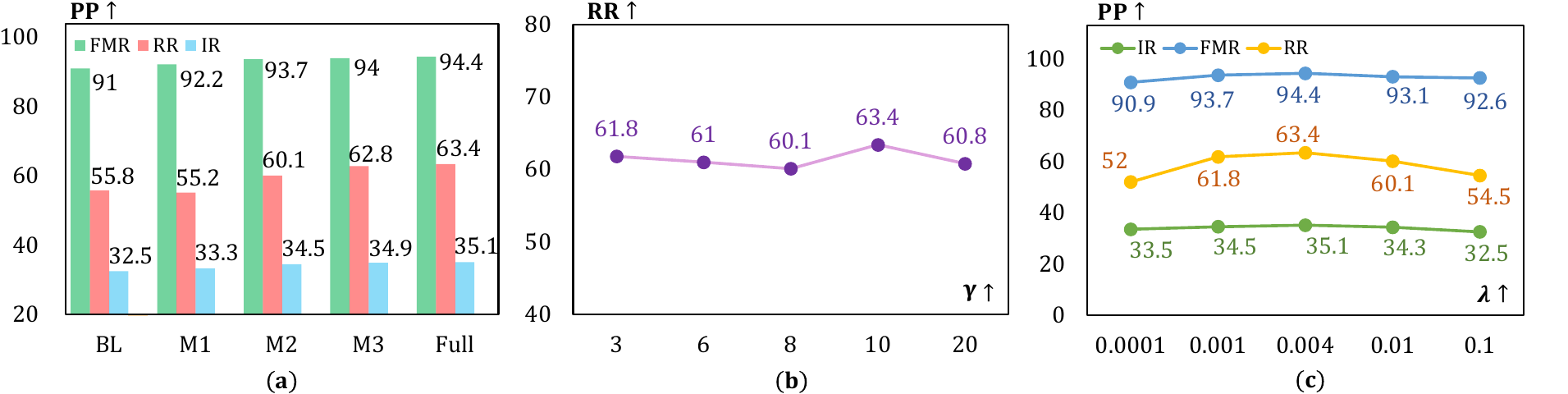}
\caption{ (a) Ablation studies of our model. (b) Ablation studies on \(\gamma\). (c) Ablation studies on \(\lambda\). }
\label{fig:fig4}
\end{figure*}

\textbf{Dataset. } \textit{RGB-D Scenes v2} consists of 14 scenes containing furniture. For each scene, we create point cloud fragments from every 25 consecutive depth frames and sample one RGB image per 25 frames. We select image-point-cloud pairs with an overlap ratio of at least 30\%. Scenes 1-8 are used for training, 9-10 for validation, and 11-14 for testing, resulting in 1,748 training pairs, 236 validation pairs, and 497 testing pairs.

The \textit{7-Scenes} is a collection of tracked RGB-D camera frames. All seven indoor scenes were recorded from a handheld Kinect RGB-D camera at 640×480 resolution. We select image-to-point-cloud pairs from each scene with at least 50\% overlap, adhering to the official sequence split for training, validation, and testing. This results in 4,048 training pairs, 1,011 validation pairs, and 2,304 testing pairs.

\textbf{Implementation Details.} We used an NVIDIA Geforce RTX 3090 GPU for training. Our Domain Classifier employs three fully connected layers with dimensions \{128, 64, 2\}. In the uncertainty estimation layer, the mean is obtained by averaging the image features at each layer, while the variance is normalized through a convolutional layer followed by a softplus activation. To prevent the variance from being zero, a small constant is added.

\textbf{Metrics.} We evaluate the models with three metrics: Inlier Ratio (IR) — the percentage of pixel-point matches with a 3D distance below 5 cm; Feature Matching Recall (FMR) — the percentage of image-point-cloud pairs with an inlier ratio above 10\%; and Registration Recall (RR) — the percentage of image-point-cloud pairs with an RMSE below 10 cm.

\begin{table}[ht]
\centering
\setlength{\tabcolsep}{1mm} 
\begin{tabular}{lcccccccc}
\toprule
\multicolumn{1}{l|}{Mtd}  & Chs & Fr & Hds & Off & Pmp & Ktn & Strs & Mean \\ \midrule
\multicolumn{1}{l|}{Mdpt} & 1.78 & 1.55 & 0.80 & 2.03 & 2.25 & 2.13 & 1.84 & 1.77 \\ \midrule
\multicolumn{9}{c}{\textit{Inlier Ratio} \(\uparrow\)} \\ \midrule
\multicolumn{1}{l|}{FC} & 34.2 & 32.8 & 14.8 & 26 & 23.3 & 22.5 & 6.0 & 22.8 \\
\multicolumn{1}{l|}{Pr} & 34.7 & 33.8 & 16.6 & 25.9 & 23.1 & 22.2 & \underline{7.5} & 23.4 \\
\multicolumn{1}{l|}{P2} & 55.2 & 46.7 & 13.0 & 36.2 & 32.0 & 32.8 & 5.8 & 31.7 \\
\multicolumn{1}{l|}{2D} & \underline{72.1} & \underline{66.0} & \underline{31.3} & \underline{60.7} & \underline{50.2} & \textbf{52.5} & \textbf{18.1} & \underline{50.1} \\
\multicolumn{1}{l|}{ours} & \textbf{73.8} & \textbf{66.7} & \textbf{33.1} & \textbf{61.7} & \textbf{50.8} & \underline{52.3} & \textbf{18.1} & \textbf{50.9} \\ \midrule
\multicolumn{9}{c}{\textit{Feature Matching Recall} \(\uparrow\)} \\ \midrule
\multicolumn{1}{l|}{FC} & \underline{99.7} & 98.2 & 69.9 & 97.1 & 83.0 & 87.7 & 16.2 & 78.8 \\
\multicolumn{1}{l|}{Pr} & 91.3 & 95.1 & \underline{76.6} & 88.6 & 79.2 & 80.6 & 31.1 & 77.5 \\
\multicolumn{1}{l|}{P2} & \textbf{100.0} & 99.3 & 58.9 & \underline{99.1} & 87.2 & 92.2 & 16.2 & 79 \\
\multicolumn{1}{l|}{2D} & \textbf{100.0} & \underline{99.6} & \textbf{98.6} & \textbf{100.0} & \underline{92.4} & \textbf{95.9} & \underline{58.2} & \underline{92.1} \\
\multicolumn{1}{l|}{ours} & \textbf{100.0} & \textbf{100.0} & \textbf{98.6} & \textbf{100.0} & \textbf{92.7} & \underline{95.6} & \textbf{64.9} & \textbf{93.1} \\ \midrule
\multicolumn{9}{c}{\textit{Registration Recall} \(\uparrow\)} \\ \midrule
\multicolumn{1}{l|}{FC} & 89.5 & 79.7 & 19.2 & 85.9 & 69.4 & 79.0 & 6.8 & 61.4 \\
\multicolumn{1}{l|}{Pr} & 69.6 & 60.7 & 17.8 & 62.9 & 56.2 & 62.6 & 9.5 & 48.5 \\
\multicolumn{1}{l|}{P2} & \underline{96.9} & 86.5 & 20.5 & 91.7 & 75.3 & 85.2 & 4.1 & 65.7 \\
\multicolumn{1}{l|}{2D} & \underline{96.9} & \textbf{90.7} & \underline{52.1} & \underline{95.5} & \underline{80.9} & \underline{86.1} & \underline{28.4} & \underline{75.8} \\
\multicolumn{1}{l|}{ours} & \textbf{98.3} & \underline{90.5} & \textbf{56.2} & \textbf{96.4} & \textbf{84.0} & \textbf{86.1} & \textbf{32.4} & \textbf{77.7} \\ \bottomrule
\end{tabular}
\caption{Evaluation results on 7Scenes. Models are abbreviated using the first two letters, and scenes are abbreviated. \textbf{Boldfaced} numbers highlight the best and the second best are \underline{underlined}.}
\end{table}

\subsection{Evaluations on Dataset}

We compared our approach with 2D3D-MATR and other baselines on the RGB-D Scenes V2 dataset (Table 1). Our method, incorporating an attention mechanism focusing on key image regions, improves inlier ratio by 2.7 percentage points (pp) and feature matching recall by 3.6 pp over previous methods. Notably, it surpasses the previous state-of-the-art in critical metric registration recall by 7 pp.

We evaluated the generalization of our method to unseen viewpoints in the 7-Scenes dataset, with results presented in Table 2. Our approach achieves an advantage of 0.8 percentage points (pp) over 2D3D-MATR in an inlier ratio. This advantage further increases to 1 pp in feature matching recall. In terms of registration recall, our method outperforms the baseline by nearly 2 pp. The 7-Scenes dataset presents more significant scale variations compared to RGB-D Scenes V2. Despite these challenges, our method demonstrates superior performance, highlighting B2-3Dnet’s adaptability to varying scales. Notably, 2D3D-MATR shows marked improvements in two challenging scenes, Heads and Stairs. In the Heads scene, the camera's proximity to surfaces amplifies small errors in 3D space on the image plane, complicating accurate correspondence extraction. The Stairs scene includes many indistinguishable repetitive patterns. Our multi-scale patch uncertainty modeling strategy effectively addresses these challenges, enabling better performance in difficult scenarios.

\subsection{Ablation Studies}

We conducted extensive ablation studies to evaluate the effectiveness of our designs on the RGB-D Scenes V2 dataset. We tested the impact of the UHMM (comprising the interaction stage and uncertainty estimation layer) and the AMAM. As illustrated in Figure 4(a), BL represents the baseline, M1 represents only the interaction stage, M2 represents only the UHMM, M3 represents the interaction stage and AMAM, and full represents all modules included. The interaction stage alone increased the inlier ratio, but without the uncertainty estimation layer, the registration recall (RR) decreased. Adding the AMAM improved all performance metrics by reducing domain differences between modalities.

In the process of selecting the total variance threshold \(\gamma\), as \(\gamma\) changed, the values exhibited some fluctuations. However, these were minor and did not significantly impact the overall results. Regarding the threshold for the coefficient \(\lambda\) of the gradient reversal layer (GRL), we found that the range between 0.001 and 0.1 is optimal, and through testing, we determined the best choice. The visual representations of these two parts are shown in Figure 4(b) and Figure 4(c).

\section{Conclusion}

We propose the B2-3Dnet, an uncertainty-aware hierarchical registration network with domain alignment. Our approach models uncertainty from hierarchically extracted image patches and interacts with point cloud features to identify key informative regions while leveraging contextual information. Additionally, we employ a domain alignment module to reduce the domain differences between images and point clouds, enhancing the accuracy and robustness of the image-to-point cloud registration task. Our method achieves state-of-the-art performance on the RGB-D Scenes V2 and 7-Scenes datasets.

\section{Acknowledgments}
This work was supported by Basic Strengthening Program Laboratory Fund (No. NKLDSE2023A009), National Natural Science Foundation of China (62394354, 62121002), Youth Innovation Promotion Association.

\clearpage


\begin{thebibliography}{40}
\providecommand{\natexlab}[1]{#1}

\bibitem[{Bolognini, Rasi, and L{\`a}davas(2005)}]{visuallocalization}
Bolognini, N.; Rasi, F.; and L{\`a}davas, E. 2005.
\newblock Visual localization of sounds.
\newblock \emph{Neuropsychologia}, 43(11): 1655--1661.

\bibitem[{Durrant-Whyte and Bailey(2006)}]{slam}
Durrant-Whyte, H.; and Bailey, T. 2006.
\newblock Simultaneous localization and mapping: part I.
\newblock \emph{IEEE robotics \& automation magazine}, 13(2): 99--110.

\bibitem[{Feng et~al.(2019)Feng, Hu, Ang, and Lee}]{2d3dmatchnet}
Feng, M.; Hu, S.; Ang, M.~H.; and Lee, G.~H. 2019.
\newblock 2d3d-matchnet: Learning to match keypoints across 2d image and 3d
  point cloud.
\newblock In \emph{2019 International Conference on Robotics and Automation
  (ICRA)}, 4790--4796. IEEE.

\bibitem[{Fischler and Bolles(1981)}]{ransac}
Fischler, M.~A.; and Bolles, R.~C. 1981.
\newblock Random sample consensus: a paradigm for model fitting with
  applications to image analysis and automated cartography.
\newblock \emph{Communications of the ACM}, 24(6): 381--395.

\bibitem[{Ganin and Lempitsky(2015)}]{grl}
Ganin, Y.; and Lempitsky, V. 2015.
\newblock Unsupervised domain adaptation by backpropagation.
\newblock In \emph{International conference on machine learning}, 1180--1189.
  PMLR.

\bibitem[{Ghanavati, Amyot, and Rifaut(2014)}]{domainunified}
Ghanavati, S.; Amyot, D.; and Rifaut, A. 2014.
\newblock Legal goal-oriented requirement language (legal GRL) for modeling
  regulations.
\newblock In \emph{Proceedings of the 6th international workshop on modeling in
  software engineering}, 1--6.

\bibitem[{Glocker et~al.(2013)Glocker, Izadi, Shotton, and Criminisi}]{7scenes}
Glocker, B.; Izadi, S.; Shotton, J.; and Criminisi, A. 2013.
\newblock Real-time RGB-D camera relocalization.
\newblock In \emph{2013 IEEE International Symposium on Mixed and Augmented
  Reality (ISMAR)}, 173--179. IEEE.

\bibitem[{He et~al.(2016)He, Zhang, Ren, and Sun}]{resnet}
He, K.; Zhang, X.; Ren, S.; and Sun, J. 2016.
\newblock Deep residual learning for image recognition.
\newblock In \emph{Proceedings of the IEEE conference on computer vision and
  pattern recognition}, 770--778.

\bibitem[{Huang et~al.(2021)Huang, Gojcic, Usvyatsov, Wieser, and
  Schindler}]{overlap}
Huang, S.; Gojcic, Z.; Usvyatsov, M.; Wieser, A.; and Schindler, K. 2021.
\newblock Predator: Registration of 3d point clouds with low overlap.
\newblock In \emph{Proceedings of the IEEE/CVF Conference on computer vision
  and pattern recognition}, 4267--4276.

\bibitem[{Kang et~al.(2023)Kang, Liao, Li, Liang, Li, Li, Dong, and
  Yang}]{cofii2p}
Kang, S.; Liao, Y.; Li, J.; Liang, F.; Li, Y.; Li, F.; Dong, Z.; and Yang, B.
  2023.
\newblock CoFiI2P: Coarse-to-Fine Correspondences for Image-to-Point Cloud
  Registration.
\newblock \emph{arXiv preprint arXiv:2309.14660}.

\bibitem[{Lai, Bo, and Fox(2014)}]{rgbdv2}
Lai, K.; Bo, L.; and Fox, D. 2014.
\newblock Unsupervised feature learning for 3d scene labeling.
\newblock In \emph{2014 IEEE International Conference on Robotics and
  Automation (ICRA)}, 3050--3057. IEEE.

\bibitem[{Lepetit, Moreno-Noguer, and Fua(2009)}]{pnp}
Lepetit, V.; Moreno-Noguer, F.; and Fua, P. 2009.
\newblock EP n P: An accurate O (n) solution to the P n P problem.
\newblock \emph{International journal of computer vision}, 81: 155--166.

\bibitem[{Li, Chen, and Lee(2018)}]{pointtonode}
Li, J.; Chen, B.~M.; and Lee, G.~H. 2018.
\newblock So-net: Self-organizing network for point cloud analysis.
\newblock In \emph{Proceedings of the IEEE conference on computer vision and
  pattern recognition}, 9397--9406.

\bibitem[{Li and Lee(2021)}]{deepi2p}
Li, J.; and Lee, G.~H. 2021.
\newblock DeepI2P: Image-to-point cloud registration via deep classification.
\newblock In \emph{Proceedings of the IEEE/CVF Conference on Computer Vision
  and Pattern Recognition}, 15960--15969.

\bibitem[{Li et~al.(2023)Li, Qin, Gao, Yi, Zhu, Guo, and Xu}]{matr2d3d}
Li, M.; Qin, Z.; Gao, Z.; Yi, R.; Zhu, C.; Guo, Y.; and Xu, K. 2023.
\newblock 2d3d-matr: 2d-3d matching transformer for detection-free registration
  between images and point clouds.
\newblock In \emph{Proceedings of the IEEE/CVF International Conference on
  Computer Vision}, 14128--14138.

\bibitem[{Lin et~al.(2017)Lin, Doll{\'a}r, Girshick, He, Hariharan, and
  Belongie}]{fpn}
Lin, T.-Y.; Doll{\'a}r, P.; Girshick, R.; He, K.; Hariharan, B.; and Belongie,
  S. 2017.
\newblock Feature pyramid networks for object detection.
\newblock In \emph{Proceedings of the IEEE conference on computer vision and
  pattern recognition}, 2117--2125.

\bibitem[{Moheimani, Vautier, and Bhikkaji(2006)}]{ppf}
Moheimani, S.~R.; Vautier, B.~J.; and Bhikkaji, B. 2006.
\newblock Experimental implementation of extended multivariable PPF control on
  an active structure.
\newblock \emph{IEEE Transactions on Control Systems Technology}, 14(3):
  443--455.

\bibitem[{Mouragnon et~al.(2006)Mouragnon, Lhuillier, Dhome, Dekeyser, and
  Sayd}]{3dreconstruction}
Mouragnon, E.; Lhuillier, M.; Dhome, M.; Dekeyser, F.; and Sayd, P. 2006.
\newblock Real time localization and 3d reconstruction.
\newblock In \emph{2006 IEEE Computer Society Conference on Computer Vision and
  Pattern Recognition (CVPR'06)}, volume~1, 363--370. IEEE.

\bibitem[{Ng and Henikoff(2003)}]{sift}
Ng, P.~C.; and Henikoff, S. 2003.
\newblock SIFT: Predicting amino acid changes that affect protein function.
\newblock \emph{Nucleic acids research}, 31(13): 3812--3814.

\bibitem[{Qi et~al.(2017)Qi, Su, Mo, and Guibas}]{pointnet}
Qi, C.~R.; Su, H.; Mo, K.; and Guibas, L.~J. 2017.
\newblock Pointnet: Deep learning on point sets for 3d classification and
  segmentation.
\newblock In \emph{Proceedings of the IEEE conference on computer vision and
  pattern recognition}, 652--660.

\bibitem[{Qin et~al.(2023)Qin, Yu, Wang, Guo, Peng, Ilic, Hu, and
  Xu}]{geotransformer}
Qin, Z.; Yu, H.; Wang, C.; Guo, Y.; Peng, Y.; Ilic, S.; Hu, D.; and Xu, K.
  2023.
\newblock Geotransformer: Fast and robust point cloud registration with
  geometric transformer.
\newblock \emph{IEEE Transactions on Pattern Analysis and Machine
  Intelligence}, 45(8): 9806--9821.

\bibitem[{Qin et~al.(2022{\natexlab{a}})Qin, Yu, Wang, Guo, Peng, and
  Xu}]{topk}
Qin, Z.; Yu, H.; Wang, C.; Guo, Y.; Peng, Y.; and Xu, K. 2022{\natexlab{a}}.
\newblock Geometric transformer for fast and robust point cloud registration.
\newblock In \emph{Proceedings of the IEEE/CVF conference on computer vision
  and pattern recognition}, 11143--11152.

\bibitem[{Qin et~al.(2022{\natexlab{b}})Qin, Yu, Wang, Guo, Peng, and
  Xu}]{circleloss1}
Qin, Z.; Yu, H.; Wang, C.; Guo, Y.; Peng, Y.; and Xu, K. 2022{\natexlab{b}}.
\newblock Geometric transformer for fast and robust point cloud registration.
\newblock In \emph{Proceedings of the IEEE/CVF conference on computer vision
  and pattern recognition}, 11143--11152.

\bibitem[{Ren et~al.(2022)Ren, Zeng, Hou, and Chen}]{corri2p}
Ren, S.; Zeng, Y.; Hou, J.; and Chen, X. 2022.
\newblock CorrI2P: Deep image-to-point cloud registration via dense
  correspondence.
\newblock \emph{IEEE Transactions on Circuits and Systems for Video
  Technology}, 33(3): 1198--1208.

\bibitem[{Rubinstein(1999)}]{crossentropy}
Rubinstein, R. 1999.
\newblock The cross-entropy method for combinatorial and continuous
  optimization.
\newblock \emph{Methodology and computing in applied probability}, 1: 127--190.

\bibitem[{Rublee et~al.(2011)Rublee, Rabaud, Konolige, and Bradski}]{orb}
Rublee, E.; Rabaud, V.; Konolige, K.; and Bradski, G. 2011.
\newblock ORB: An efficient alternative to SIFT or SURF.
\newblock In \emph{2011 International conference on computer vision},
  2564--2571. Ieee.

\bibitem[{Rusu, Blodow, and Beetz(2009)}]{fpfh}
Rusu, R.~B.; Blodow, N.; and Beetz, M. 2009.
\newblock Fast point feature histograms (FPFH) for 3D registration.
\newblock In \emph{2009 IEEE international conference on robotics and
  automation}, 3212--3217. IEEE.

\bibitem[{Sarlin et~al.(2020)Sarlin, DeTone, Malisiewicz, and
  Rabinovich}]{superglue}
Sarlin, P.-E.; DeTone, D.; Malisiewicz, T.; and Rabinovich, A. 2020.
\newblock Superglue: Learning feature matching with graph neural networks.
\newblock In \emph{Proceedings of the IEEE/CVF conference on computer vision
  and pattern recognition}, 4938--4947.

\bibitem[{Sontag(1998)}]{iss}
Sontag, E.~D. 1998.
\newblock Comments on integral variants of ISS.
\newblock \emph{Systems \& Control Letters}, 34(1-2): 93--100.

\bibitem[{Strehl, Ghosh, and Mooney(2000)}]{cos}
Strehl, A.; Ghosh, J.; and Mooney, R. 2000.
\newblock Impact of similarity measures on web-page clustering.
\newblock In \emph{Workshop on artificial intelligence for web search (AAAI
  2000)}, volume~58, 64.

\bibitem[{Sun et~al.(2021)Sun, Shen, Wang, Bao, and Zhou}]{loftr}
Sun, J.; Shen, Z.; Wang, Y.; Bao, H.; and Zhou, X. 2021.
\newblock LoFTR: Detector-free local feature matching with transformers.
\newblock In \emph{Proceedings of the IEEE/CVF conference on computer vision
  and pattern recognition}, 8922--8931.

\bibitem[{Sun et~al.(2020)Sun, Cheng, Zhang, Zhang, Zheng, Wang, and
  Wei}]{circleloss}
Sun, Y.; Cheng, C.; Zhang, Y.; Zhang, C.; Zheng, L.; Wang, Z.; and Wei, Y.
  2020.
\newblock Circle loss: A unified perspective of pair similarity optimization.
\newblock In \emph{Proceedings of the IEEE/CVF conference on computer vision
  and pattern recognition}, 6398--6407.

\bibitem[{Thomas et~al.(2019)Thomas, Qi, Deschaud, Marcotegui, Goulette, and
  Guibas}]{kpfcnn}
Thomas, H.; Qi, C.~R.; Deschaud, J.-E.; Marcotegui, B.; Goulette, F.; and
  Guibas, L.~J. 2019.
\newblock Kpconv: Flexible and deformable convolution for point clouds.
\newblock In \emph{Proceedings of the IEEE/CVF international conference on
  computer vision}, 6411--6420.

\bibitem[{Van~der Maaten and Hinton(2008)}]{tsne}
Van~der Maaten, L.; and Hinton, G. 2008.
\newblock Visualizing data using t-SNE.
\newblock \emph{Journal of machine learning research}, 9(11).

\bibitem[{Vaswani et~al.(2017)Vaswani, Shazeer, Parmar, Uszkoreit, Jones,
  Gomez, Kaiser, and Polosukhin}]{transformer}
Vaswani, A.; Shazeer, N.; Parmar, N.; Uszkoreit, J.; Jones, L.; Gomez, A.~N.;
  Kaiser, {\L}.; and Polosukhin, I. 2017.
\newblock Attention is all you need.
\newblock \emph{Advances in neural information processing systems}, 30.

\bibitem[{Wang et~al.(2021)Wang, Chen, Cui, Qin, Lu, Yu, Zhao, Dong, Zhu,
  Trigoni et~al.}]{p2}
Wang, B.; Chen, C.; Cui, Z.; Qin, J.; Lu, C.~X.; Yu, Z.; Zhao, P.; Dong, Z.;
  Zhu, F.; Trigoni, N.; et~al. 2021.
\newblock P2-net: Joint description and detection of local features for pixel
  and point matching.
\newblock In \emph{Proceedings of the IEEE/CVF International Conference on
  Computer Vision}, 16004--16013.

\bibitem[{Wang et~al.(2024)Wang, He, Peng, Tan, and Zhou}]{efficientloftr}
Wang, Y.; He, X.; Peng, S.; Tan, D.; and Zhou, X. 2024.
\newblock Efficient LoFTR: Semi-dense local feature matching with sparse-like
  speed.
\newblock In \emph{Proceedings of the IEEE/CVF Conference on Computer Vision
  and Pattern Recognition}, 21666--21675.

\bibitem[{Yu et~al.(2021)Yu, Li, Saleh, Busam, and Ilic}]{cofinet}
Yu, H.; Li, F.; Saleh, M.; Busam, B.; and Ilic, S. 2021.
\newblock Cofinet: Reliable coarse-to-fine correspondences for robust
  pointcloud registration.
\newblock \emph{Advances in Neural Information Processing Systems}, 34:
  23872--23884.

\bibitem[{Yu et~al.(2019)Yu, Li, Yang, Hospedales, and Xiang}]{uncertainty}
Yu, T.; Li, D.; Yang, Y.; Hospedales, T.~M.; and Xiang, T. 2019.
\newblock Robust person re-identification by modelling feature uncertainty.
\newblock In \emph{Proceedings of the IEEE/CVF international conference on
  computer vision}, 552--561.

\bibitem[{Zhang et~al.(2022)Zhang, Zhou, Sun, and Jha}]{lightweightcnn}
Zhang, X.; Zhou, J.; Sun, W.; and Jha, S.~K. 2022.
\newblock A Lightweight CNN Based on Transfer Learning for COVID-19 Diagnosis.
\newblock \emph{Computers, Materials \& Continua}, 72(1).

\end{thebibliography}
\end{document}